\ifdefined\pdftexversion\pdfoutput=1\fi 
\newif\ificml
\icmlfalse

\documentclass{article}

\usepackage{microtype}
\usepackage{graphicx}
\usepackage{subcaption}
\usepackage{booktabs}
\usepackage{hyperref}

\ificml
  \usepackage{icml2026}            
\else
  \usepackage[preprint]{icml2026}  
  \hypersetup{pdftitle={Diversity over Scale: Recursive Self-Improvement with Ensembles of Small Open-Weight Models},
    pdfsubject={}}
\fi

\usepackage{amsmath}
\usepackage{amssymb}
\usepackage{mathtools}
\usepackage[capitalize,noabbrev]{cleveref}
\usepackage{multirow}
\usepackage{enumitem}
\usepackage{iftex}
\ifPDFTeX
  \usepackage[T1]{fontenc}  
\fi
\usepackage{tikz}
\usetikzlibrary{arrows.meta,shapes.geometric,decorations.pathreplacing,decorations.markings,shadows}

\newcommand{\FailuresShown}{5}
\newcommand{\TemplateMarkers}{3}
\newcommand{\PatchLines}{293}
\newcommand{\ConfigValues}{6}
\newcommand{\OELines}{8,893}

\newcommand{\nLite}{22}

\newcommand{\EnsGainTest}{38.1}

\newcommand{\StallSingleMin}{80}
\newcommand{\StallSingleMax}{99}

\newcommand{\ValidMemberMin}{8}
\newcommand{\ValidMemberMax}{70}

\newcommand{\CmpEnsVsLlamaLMedian}{16.7}

\newcommand{\EnsVsMembersMedian}{25.4}

\newcommand{\EnsGapQwenLMedian}{7.1}

\newcommand{\BestHindGainTest}{32.3}

\newcommand{\EnsBeatsLlamaL}{19}

\newcommand{\IdeaGainGran}{31}

\newcommand{\IdeaGainLlamaS}{12}

\newcommand{\StartMedals}{2}

\newcommand{\EnsCostShareQwenL}{75}

\newcommand{\TeaserReachPct}{63}

\newcommand{\HandoffShare}{59}

\newcommand{\ExMini}{0.429}
\newcommand{\ExGran}{0.524}

\newcommand{\ExLlamaS}{0.542}

\newcommand{\QualSmallImproved}{70}
\newcommand{\QualSmallFamilies}{0.97}

\newcommand{\QualEnsFamilies}{1.14}
\newcommand{\QualEnsTrees}{45}
\newcommand{\QualLargeImproved}{82}

\newcommand{\QualLargeTrees}{75}

\newcommand{\QualStartTrees}{32}
\newcommand{\QualSmallN}{110}
\newcommand{\QualLargeN}{44}
\newcommand{\KernStart}{147,734}

\definecolor{qwen}{HTML}{43A36B}
\definecolor{qwenbig}{HTML}{17704A}
\definecolor{llama}{HTML}{5B8FD9}
\definecolor{llamabig}{HTML}{24589E}
\definecolor{ministral}{HTML}{EB6834}
\definecolor{granite}{HTML}{6250D6}
\definecolor{gemma}{HTML}{C2447A}
\definecolor{amberline}{HTML}{D98E04}
\definecolor{amberfill}{HTML}{FFF3D6}

\newcommand{\method}{DiVE}
\setlist{itemsep=1pt,topsep=3pt,leftmargin=12pt}
\icmltitlerunning{Diversity over Scale: RSI with Small Open-Weight Ensembles}

\begin{document}

\twocolumn[
  \icmltitle{Diversity over Scale: Recursive Self-Improvement\\with Ensembles of Small Open-Weight Models}

  \begin{icmlauthorlist}
    \icmlauthor{Navapat Nananukul}{inst}
    \icmlauthor{Mayank Kejriwal}{inst}
  \end{icmlauthorlist}

  \vskip 0.1in
  {\centering\small Information Sciences Institute, University of Southern California\par}

  \icmlaffiliation{inst}{Information Sciences Institute, University of Southern California}
  \icmlcorrespondingauthor{Navapat Nananukul}{nananuku@usc.edu}

  \icmlkeywords{recursive self-improvement, autoresearch, LLM ensembles, open-weight models, MLE-bench, evolutionary code search}

  \vskip 0.3in
]

\begin{abstract}
The advancement of large language models (LLMs) has significantly enhanced the capabilities of agents across various tasks. Recursive self-improvement (RSI) loops, in which a language model repeatedly rewrites a program and builds on its previous
version, now perform well in machine-learning tasks, but they rely on frontier models and hundreds of calls per run. Small open-weight models are far cheaper, yet when used alone, their programs often crash and their improvements stall. We propose \method{}
(\emph{diverse models evolve}), an ensemble method for RSI that aims to improve creativity and diversity of generated solutions. \method{} uses three components: a \emph{shared memory}, a \emph{failure log}, and \emph{best-idea-wins rounds}. This allows models to write candidate solutions from the same context and retain the best idea at each iteration. On \nLite{} MLE-bench Lite competitions, an ensemble of five 8--12B open-weight models has
\EnsVsMembersMedian\% less test error than its average member and \CmpEnsVsLlamaLMedian\% less than Llama~3.3~70B, and
comes within \EnsGapQwenLMedian\% of Qwen~3.5~122B-A10B at \EnsCostShareQwenL\% of its cost. Diversity among small models can thus stand in for part of the scale that RSI relies on.
\end{abstract}

\begin{figure}[t]
  \centering
  \includegraphics[width=\columnwidth]{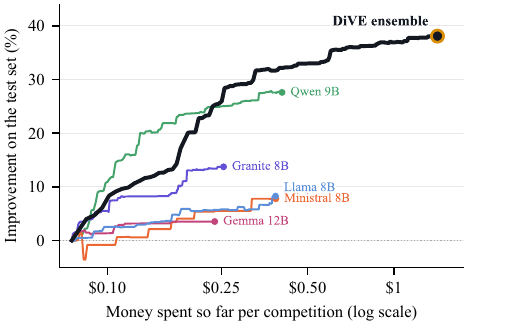}
  \caption{\textbf{\method{} matches the best small model at \TeaserReachPct\% of its cost and ends 25\%
  above the average of five individual runs.} Mean improvement over the shared start on the test set against money spent, over \nLite{} MLE-bench Lite competitions; costs include the zero-shot rounds that chose the start.}
  \label{fig:teaser}
\end{figure}

\section{Introduction}
\label{sec:intro}

An autoresearch loop is a sequence of experiments linked by parent--child relations: a model picks a program from an earlier iteration, proposes a change, and an evaluator scores the result~\citep{romeraparedes2024funsearch,novikov2025alphaevolve,sharma2025openevolve,lange2025shinkaevolve}. When the
model also rewrites the notes that steer its next proposals, the loop improves both what it produces and how it produces it. We call this \emph{recursive self-improvement} (RSI), in the inference-time sense of recent
self-improving agents~\citep{zelikman2023stop,fernando2023promptbreeder,zhang2025dgm}: no weights change
(\cref{sec:formulation}). On MLE-bench~\citep{chan2024mlebench}, the best such agents now medal in up to 80\% of
the Lite competitions, with frontier models run for 12--24 hours per attempt \citep{chan2024mlebench}.

Scale alone does not keep these runs moving. Every iteration starts from the current best program, so a single model
tends to refine the idea it already has, and after the early gains the best score stops rising. In our runs, a
single model goes at least 20 rounds without a new best for \StallSingleMin--\StallSingleMax\% of its rounds
(\cref{sec:bestidea}), and each attempt in one direction adds context that pulls the next attempt the same
way. Existing methods try to solve this by drawing more diversity from one model, through verbalized sampling~\citep{zhang2025verbalized} or through resets that discard the conversation~\citep{cai2026forkflush,tan2026forgetting}. All of
them sample more widely from \emph{one} model's distribution, which post-training has already narrowed~\citep{zhang2025verbalized,jiang2025hivemind}.

\begin{figure*}[!t]
  \centering
  \resizebox{\textwidth}{!}{\input{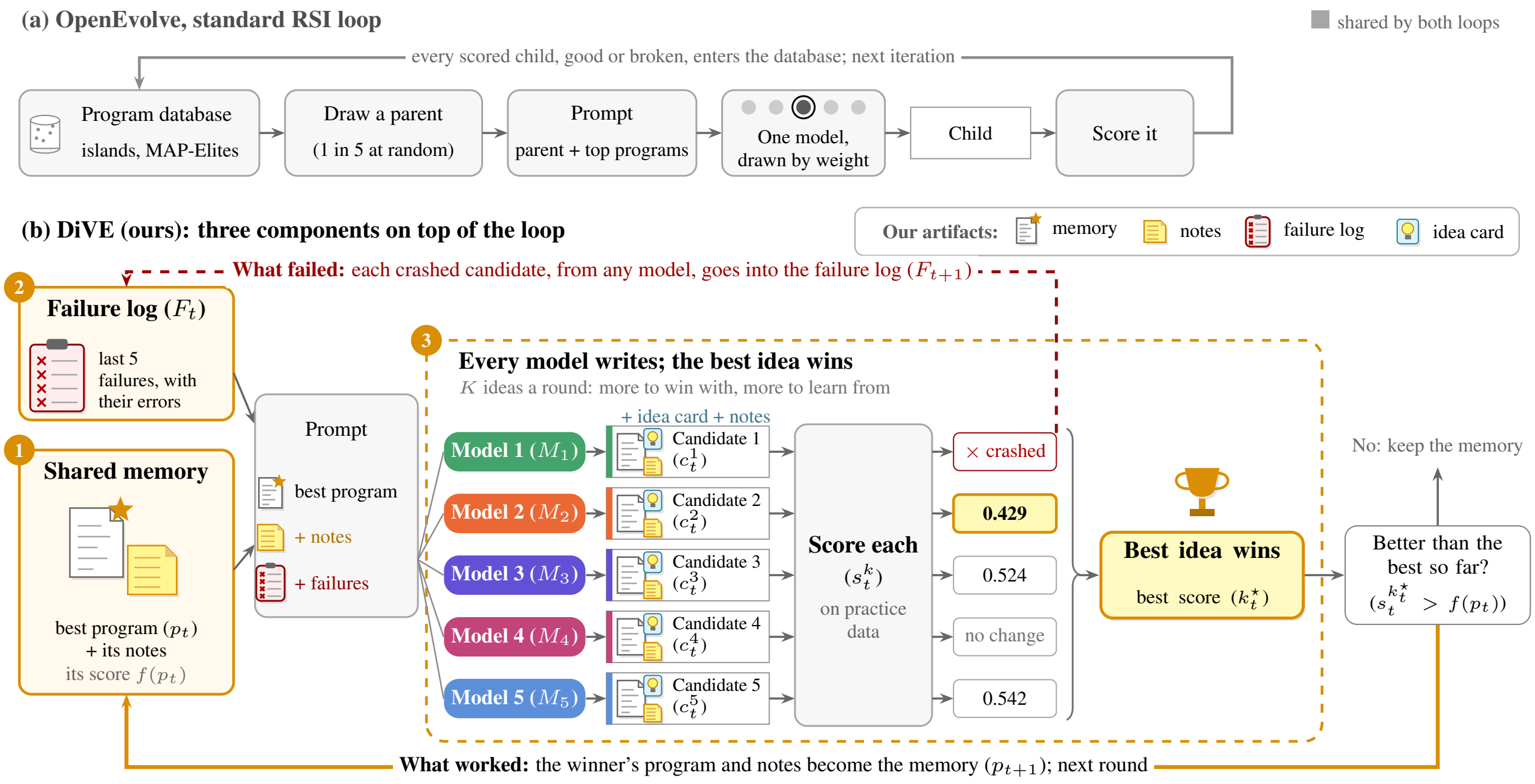}}
  \caption{A comparison of simplified OpenEvolve and \method{}. (a) OpenEvolve, the standard RSI loop, draws one model per iteration by a fixed weight and adds every scored child to a program database from which the next parent is drawn.
  (b) \method{} adds three components. (1)~\emph{Shared memory}, (2)~\emph{Failure log}, and (3)~\emph{Best idea
  wins}: all $K$ models write a candidate from the same prompt, and the best-scoring one replaces
  the memory if it beats it. Components 1 and 2 are part of every run in our study, single-model or ensemble. Scores
  are from a real ML evaluator at each round.}
  \label{fig:method}
\end{figure*}

Several models are not new to this search, but existing systems, including OpenEvolve, the standard open-source RSI
loop (\cref{fig:method}a), use them to trade cost against quality: they choose one model per proposal \emph{before} it writes~\citep{novikov2025alphaevolve,sharma2025openevolve,codeevolve2025,lange2025shinkaevolve}. A weak model's turn is
thus spent whether or not its program runs. Whether several small models can stand in for one large model has not been tested.

In this work, we use several models for diversity. Models from different providers (Alibaba, Mistral AI, IBM, Google, Meta), trained on different data, bring different priors, coding habits and failure modes.
Alone, each is a poor researcher: only \ValidMemberMin--\ValidMemberMax\% of their programs run, and the best of them
changes from competition to competition. We propose \method{} (\emph{diverse models evolve}), which turns them into one
researcher with three components (\cref{fig:method}b): a \emph{shared memory}, the best program so far with its author's notes, on which every model builds; a \emph{failure log}, every model's recent crashes with their errors,
shown in every prompt; and \emph{best-idea-wins rounds}, in which every model writes a candidate from the same context. Then an evaluator scores all candidates and selects the best program as the winner.

Choosing after writing changes what a weak model costs. A model whose programs rarely run costs one cheap call per round instead of a share of the run, one model's crash warns all of them, and the five calls of a round cost less than
one call to a 122B model. Small open-weight models also matter beyond price: frozen weights keep runs reproducible
after an API is retired, they can be self-hosted for private data, and they fit on one consumer
GPU~\citep{belcak2025slm}.

We ask whether such an ensemble matches a large model, and whether the gain is the ensemble's or only its best member's. Every competition gets eight runs that share the start program, prompt, memory, failure log, evaluator and
budget: the ensemble, each of its five members alone, and the two larger models, Llama~3.3~70B and
Qwen~3.5~122B-A10B.

\paragraph{Contributions.}
\begin{itemize}
  \item \textbf{A new ensemble method for RSI.} We propose \method{}, in which several models improve one program through a shared memory, a failure log  and best-idea-wins rounds.
  \item \textbf{Empirical results show that an \method{} ensemble of small models can beat larger models at lower costs.} Five 8--12B open-weight models in \method{} beat
  Llama~3.3~70B in \EnsBeatsLlamaL{} of \nLite{} MLE-bench Lite competitions and come within \EnsGapQwenLMedian\% of
  Qwen~3.5~122B-A10B at \EnsCostShareQwenL\% of its cost.
  \item \textbf{\method{} ensemble reduces the number of crashes of small models}. \method{} removes \EnsGainTest\% of the start's
  test error, against \BestHindGainTest\% for the best member picked in hindsight; every member contributes
  \IdeaGainLlamaS--\IdeaGainGran\% of the improvement, and \HandoffShare\% of improvements build on another member's
  program.
\end{itemize}

\section{Problem formulation}
\label{sec:formulation}

\paragraph{Program search.} The input is a task $\tau$: a problem description, its data and a score $f$. For a
program $p$, $f(p)$ runs $p$ on a \emph{practice split} of the data and returns its score, or $-\infty$ if $p$
\emph{fails}, that is, crashes or breaks a resource limit. A language model writes programs from a \emph{context}: the
task $\tau$, a \emph{parent} program, and the \emph{strategy notes} written with that parent, at most 150 words on what
to try, what to avoid and how to search. Each reply is a new program with rewritten notes. The model's weights stay
fixed, so the search runs over its context: starting from a program $p_1$, each of $T$ rounds scores the new program
and keeps it, with its notes, as the next parent only if it beats the current one. Better programs and notes thus
shape every later context, and the search improves how it searches along with what it finds; this is the RSI we study.
After $T$ rounds, the search returns its best program, which is graded once on a test set it never sees.

\paragraph{The \method{} round.} \method{} runs $K$ language models $M_1,\dots,M_K$ ($K=1$ for a single model) that
share two objects. The \emph{shared memory} $p_t$, the best program before round $t$ with its notes, is the parent of
every candidate in round $t$. The \emph{failure log} $F_t$ holds the $L$ newest failed candidates, each with its idea
and error; $F_1$ is empty. In round $t$, every model $M_k$ writes a candidate $c^k_t$ from one prompt holding the task,
$p_t$ and $F_t$; the best-scoring candidate replaces $p_t$ if it scores higher:
\begingroup\allowdisplaybreaks
\begin{align}
c^k_t &\sim M_k(\cdot \mid p_t, F_t), \qquad s^k_t = f(c^k_t), \label{eq:propose}\\
k^\star_t &= \arg\max\nolimits_{k} s^k_t, \label{eq:select}\\
p_{t+1} &= \begin{cases} c^{k^\star_t}_t & \text{if } s^{k^\star_t}_t > f(p_t),\\
p_t & \text{otherwise,}\end{cases} \label{eq:accept}\\
F_{t+1} &= \text{the $L$ newest of } F_t \cup \{c^k_t : s^k_t = -\infty\}. \label{eq:fail}
\end{align}
\endgroup

\section{Method: \method{}}
\label{sec:method}

\method{} adds three components to the RSI loop (\cref{fig:method}b, \cref{alg:ens}). Each answers one way small models
fail: most of their programs crash (\cref{sec:memory}), a crash teaches only the model that wrote it
(\cref{sec:faillog}), and a single model stalls (\cref{sec:bestidea}). \Cref{sec:together} explains how the three
reinforce one another.

\begin{algorithm}[t]
\caption{\method{}: one run of best-of-$K$ rounds}
\label{alg:ens}
\begin{algorithmic}[1]
\REQUIRE models $M_1,\dots,M_K$; start program $p_1$; score $f$; rounds $T$; log size $L$
\STATE $F_1 \gets \emptyset$ \COMMENT{failure log}
\FOR{$t = 1, \dots, T$}
  \FORALL{$k = 1, \dots, K$ \textbf{in parallel}}
    \STATE $c^k_t \sim M_k(\cdot \mid p_t, F_t)$;\; $s^k_t \gets f(c^k_t)$
  \ENDFOR
  \STATE $k^\star_t \gets \arg\max_k s^k_t$ \COMMENT{the best idea wins}
  \IF{$s^{k^\star_t}_t > f(p_t)$}
    \STATE $p_{t+1} \gets c^{k^\star_t}_t$ \COMMENT{shared memory}
  \ELSE
    \STATE $p_{t+1} \gets p_t$
  \ENDIF
  \STATE $F_{t+1} \gets$ $L$ newest of $F_t \cup \{c^k_t : s^k_t = -\infty\}$
\ENDFOR
\STATE \textbf{return} $p_{T+1}$ \COMMENT{graded once on the test set}
\end{algorithmic}
\end{algorithm}

\subsection{Shared memory}
\label{sec:memory}

A shared memory is the main component of the RSI context in every round. Standard RSI methods usually keep only one
model's own, so each proposal extends that model's previous ideas. When ensembling the generated candidates, we
keep the best program by its score, together with the notes its author wrote to maximize the quality of that context given a set of LLMs. This program, $p_t$, is the parent of every model in the next round, so one model's idea is built on by all the others. A
candidate enters the memory only if it beats $p_t$ (\cref{eq:accept}), so the memory never holds a broken or worse
program. The notes travel with the program: every prompt ends with them and a request to rewrite them, and the
winner's rewritten notes replace the old ones.

\subsection{Failure log}
\label{sec:faillog}

Small language models tend to write programs that crash, especially when they attempt complex machine-learning
algorithms. We use these failures as a signal instead of discarding them. The failure log $F_t$ holds
the $L$ newest failed candidates from any model (\cref{eq:fail}), each with its idea and the error and the program line
where it crashed or was stopped (time, memory or invalid output), and every prompt includes it. One model's crash thus warns every model in the next round, and the ensemble does not repeat the same mistake. To make each failure readable as an idea, every candidate must open with a short \emph{idea card} that names its idea, says why it should work and lists its steps; a program without one is rejected before it runs. Failures never enter the shared memory; they only inform the next attempt.

\subsection{Best idea wins}
\label{sec:bestidea}

Models trained by different developers succeed and fail on different problems, and a single model tends to stall on
its own idea. Best-idea-wins rounds turn this diversity into progress. Each round builds one context and sends it to
all $K$ models in parallel; every model writes a candidate, all $K$ candidates are scored, and the best one wins
(\cref{eq:propose,eq:select}). The winner is chosen after writing, by its measured score, not by a weight fixed in
advance, so the best idea of the round wins whichever model had it, and a weak model costs one call per round rather
than a share of the run. A round costs $K$ calls and $K$ evaluations and takes as long as the slowest model; with small models, each call stays cheap.

\subsection{How the components work together}
\label{sec:together}

The three components complement one another. Best-idea-wins rounds give the memory $K$ ideas to choose from and the
failure log $K$ chances to learn from; the shared memory passes the winner's program and notes to every model; and the failure log passes every model's mistakes to all the others. The ensemble thus behaves as one researcher that remembers what worked, warns itself about what failed, and keeps whichever idea scores best.

\section{Experimental setup}
\label{sec:setup}

Our experiments answer four research questions. \textbf{RQ1}: does the ensemble improve a program over each of
its members? \textbf{RQ2}: does the ensemble of five small models together match the performance of larger models, and how does that compare with scaling one model up (\cref{sec:vs-large,sec:scale})? \textbf{RQ3}: how much does the ensemble cost to reach a larger model's result
(\cref{sec:cost})? \textbf{RQ4}: does the ensemble's advantage carry over to kernel optimization, a task outside
machine-learning engineering (\cref{sec:kernel})?

\begin{table}[t]
\centering\footnotesize
\caption{\textbf{Models.} In ensemble: whether the model is one of the ensemble's five members.}
\label{tab:models}
\setlength{\tabcolsep}{3pt}
\resizebox{\columnwidth}{!}{\begin{tabular}{@{}llll@{}}
\toprule
Model & Developer & Params & In ensemble \\
\midrule
Qwen 3.5 9B~\citep{qwen2026qwen35} & Alibaba & 9B & Y \\
Ministral 8B (2512)~\citep{mistral2025ministral3} & Mistral AI & 8B & Y \\
Granite 4.2 8B~\citep{ibm2026granite42} & IBM & 8B & Y \\
Gemma 3 12B~\citep{gemma2025gemma3} & Google & 12B & Y \\
Llama 3.1 8B~\citep{grattafiori2024llama3} & Meta & 8B & Y \\
\midrule
Llama 3.3 70B & Meta & 70B & N \\
Qwen 3.5 122B-A10B & Alibaba & 122B & N \\
\bottomrule
\end{tabular}}
\end{table}

\paragraph{Tasks.} \emph{MLE-bench Lite}~\citep{chan2024mlebench} is MLE-bench's split of \nLite{} Kaggle
machine-learning competitions: image classification, text classification, tabular prediction, audio
classification, text normalization and image denoising. We run all of them with MLE-bench's data
preparation, grader and medal thresholds (\cref{app:competitions}).

\paragraph{Models and runs.} \Cref{tab:models} lists the five members and the two large siblings.
Qwen~3.5~122B-A10B is a mixture of experts with about 10B active parameters, so we compare models by cost, not size.
All models are open-weight and called through OpenRouter providers (\cref{app:prompt}), with up to 8,192 output tokens.
Each competition gets eight runs: the ensemble, each member alone, Llama~3.3~70B and Qwen~3.5~122B-A10B. Each run lasts 200 rounds of \cref{eq:propose,eq:select,eq:accept,eq:fail}, with $K=1$ for a single model (200 programs) and $K=5$ for
the ensemble (1,000 programs), seed 401, and the same prompt, shared memory, failure log, evaluator and start. Every run
thus has components 1 and 2, and the comparisons isolate the ensemble. All runs used one 8-core laptop with 16\,GB and
no GPU.

\paragraph{Evaluator and start program.} During the search,
$f$ is MLE-bench's grader on the practice split, MLE-bench's public training data split again with the competition's
own split code. The returned program is run on the full public data and graded once on MLE-bench's test set. Before the runs, each model gets up to 30 calls to write a runnable program $p_1$.

\paragraph{Metrics.} With error $e = 1-s$ for a higher-is-better score $s$ and $e=s$ otherwise, a run's
\emph{improvement over the start} is $(e_{\text{start}} - e_{\text{final}})/e_{\text{start}}$, which puts competitions
with different metrics on one scale; the \emph{relative gain} of run $b$ over run $a$ is $(e_a - e_b)/e_a$. We also
report MLE-bench medals, the share of the Kaggle leaderboard beaten, the share of a run's programs that are
\emph{valid} (run without failing), and all-in API cost. Means and medians are over competitions; win counts get a
two-sided sign test.

\section{Results on MLE-bench Lite}
\label{sec:results}

\begin{table*}[t]
\centering
\caption{\textbf{Results of each run, averaged over 22 MLE-bench Lite competitions.} Improvement over the start: the
share of the start program's error that the final program removes, on the practice split and the test set. Test vs.\
own ZS: the same, measured from the model's own zero-shot program. Medals and $\geq$Median: competitions with a medal,
or with a medal or a place above the median team. Valid: share of the run's programs that ran. All-in cost: API cost
per competition.}
\label{tab:per-run}
\resizebox{\textwidth}{!}{\begin{tabular}{lrrrrrrrr}
\toprule
 & \multicolumn{2}{c}{Improvement over the start (\%)} & Test vs. & \multicolumn{3}{c}{Leaderboard (test)} & & All-in cost \\
\cmidrule(lr){2-3}\cmidrule(lr){5-7}
Run & Practice split & Test set & own ZS (\%) & Medals & $\geq$Median & Teams beaten & Valid & per comp. \\
\midrule
\textbf{Ensemble (5 small)} & \textbf{41.7} & \textbf{38.1} & \textbf{--} & \textbf{4} & \textbf{9} & \textbf{45} & \textbf{38} & \textbf{\$1.42} \\
\midrule
Qwen 3.5 9B & 28.5 & 27.6 & 35.5 & 3 & 5 & 38 & 27 & \$0.41 \\
Ministral 8B & 13.9 & 7.8 & 21.6 & 3 & 4 & 29 & 8 & \$0.39 \\
Granite 4.2 8B & 15.4 & 13.7 & 25.1 & 3 & 3 & 31 & 70 & \$0.25 \\
Gemma 3 12B & 3.6 & 3.5 & 19.9 & 2 & 2 & 26 & 69 & \$0.24 \\
Llama 3.1 8B & 10.5 & 8.3 & 27.1 & 2 & 2 & 28 & 39 & \$0.39 \\
\midrule
Llama 3.3 70B & 14.6 & 12.4 & 26.3 & 2 & 2 & 30 & 66 & \$0.58 \\
Qwen 3.5 122B-A10B & 41.9 & 41.4 & 37.4 & 6 & 9 & 49 & 82 & \$1.90 \\
\bottomrule
\end{tabular}
}
\end{table*}

\begin{figure*}[t]
  \centering
  \includegraphics[width=\textwidth]{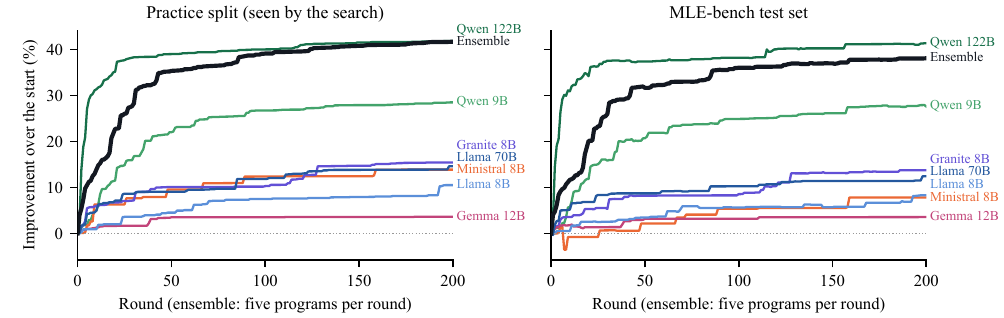}
  \caption{\textbf{Improvement over the start program, round by round,} on the practice split (left) and the test set
  (right), averaged over 22 competitions. Each curve follows the program its run would return at that round; an
  ensemble round writes five programs.}
  \label{fig:curves}
\end{figure*}

\subsection{The ensemble beats the average small models by 25.4\%}
\label{sec:members}

\begin{table}[t]
\centering\footnotesize
\caption{\textbf{What each small model contributes inside the ensemble.} New bests:
\% of the ensemble's new best programs that it wrote. Gain: its share of the ensemble's total improvement (\%). Final:
competitions where it wrote the ensemble's final program. Alone: competitions where it was the best of small models.}
\label{tab:whose-ideas}
\setlength{\tabcolsep}{3pt}
\begin{tabular}{@{}lrrrr@{}}
\toprule
 & \multicolumn{2}{c}{Share (\%)} & \multicolumn{2}{c}{Competitions} \\
\cmidrule(lr){2-3}\cmidrule(lr){4-5}
Member & New bests & Gain & Final & Alone \\
\midrule
Granite 8B & 26 & 31 & 4 & 3 \\
Ministral 8B & 8 & 21 & 1 & 3 \\
Qwen 9B & 14 & 21 & 3 & 12 \\
Gemma 12B & 26 & 16 & 5 & 1 \\
Llama 8B & 25 & 12 & 9 & 3 \\
\bottomrule
\end{tabular}
\end{table}

\Cref{tab:per-run} compares the ensemble with its five small models, each run alone from the same start program. The
ensemble's final program is 25.4\% better than the average small model's (median over competitions) and better in 20
of 21 competitions ($p<0.001$). It improves on the start program by 38.1\%, against 12.2\% for the average small model,
and finds a new best program 14.1 times per competition, against 0.8 to 5.4 times for the small models alone
(\cref{tab:cost-per-gain}). \emph{Against the best small model}, Qwen~3.5~9B, the ensemble is still 5.8\% better and
better in 13 of 20 competitions ($p=0.26$). Against the best small model picked per competition after the runs, it is
6.1\% better and better in 12 of 21 ($p=0.66$). Neither difference is significant: the ensemble matches the best small
model without knowing in advance which one it is.

\subsection{Five \texorpdfstring{8--12B}{8-12B} models beat a 70B model by 16.7\% and come within 7.1\% of a 122B
model}
\label{sec:vs-large}

The ensemble's final program is 16.7\% better than Llama~3.3~70B's (median) and better in 19 of 21 competitions
($p<0.001$). It is 7.1\% worse than Qwen~3.5~122B-A10B's and better in only 5 of 21 ($p=0.03$). From the same start
program, the ensemble improves by 38.1\%, Llama~3.3~70B by 12.4\% and Qwen~3.5~122B-A10B by 41.4\%. The ensemble finds
a new best 14.1 times per competition, against 2.8 for Llama~3.3~70B and 9.0 for Qwen~3.5~122B-A10B
(\cref{tab:cost-per-gain}). It climbs more slowly than the 122B model in the first rounds but keeps improving for
longer (\cref{fig:curves}). Per program written, it finds new bests less often than Qwen~3.5~122B-A10B (1.4 against
4.5 per 100 programs); its advantage comes from writing five cheap programs per round.

\subsection{A larger model lowers Qwen's error by 8.9\%}
\label{sec:scale}

Scaling up within a family helps Qwen but not Llama. Qwen~3.5~122B-A10B's final program is 8.9\% better than
Qwen~3.5~9B's (median) and better in 15 of 20 competitions ($p=0.04$). Llama~3.3~70B's is no better than
Llama~3.1~8B's (median 0.0\%, better in 10 of 21, $p=1.00$), although more of its programs run (66\% against 39\%).
RSI helps models of both sizes: 117 of 170 runs improve on the start program, by 18.9\% on average. Scale and RSI
together take Qwen~3.5~9B's own zero-shot program to Qwen~3.5~122B-A10B's final program, a 48.2\% improvement. The
ensemble's final program is 30.3\% better than Llama~3.3~70B's on average, six times the 70B model's 5.0\% gain over
Llama~3.1~8B.

\subsection{The ensemble matches the 70B model in 21 of 22 competitions at
\texorpdfstring{5.2$\times$}{5.2x} lower cost}
\label{sec:cost}

A full run costs \$1.42 per competition for the ensemble, \$0.58 for Llama~3.3~70B and \$1.90 for
Qwen~3.5~122B-A10B. The ensemble reaches Llama~3.3~70B's final score in 21 of 22 competitions, at a median of round 1,
for a median \$0.10 against \$0.49 for the 70B model's full run (5.2$\times$ less). In 9 of these competitions the
start program already matches it. The ensemble reaches Qwen~3.5~122B-A10B's final score in 5 competitions, at a median
of round 18, for \$0.18 against \$1.76 (7.3$\times$ less). \Cref{tab:leaderboard} places our runs next to MLE-bench
leaderboard agents, which use frontier models and 12--24 hours per competition.

\section{Results on a kernel-optimization task}
\label{sec:kernel}

We ran the same setting for 200 rounds on Anthropic's performance take-home~\citep{anthropic2026takehome}, which
asks for a faster kernel on a simulated VLIW machine, scored in cycles. All runs start from the take-home's kernel
(147,734 cycles). We test two kinds of feedback: \emph{bare}, which
only says that a wrong kernel is wrong, and \emph{debug}, which names the first failing check log (\cref{tab:kernel},
\cref{app:kernel}).

\textbf{The ensemble is the only run that improves the kernel under both kinds of feedback.} It removes 11.1\% of the
cycles with bare feedback and 8.3\% with debug feedback. None of the five small models improves the kernel alone, in
either setting: their best kernel is 10 cycles faster than the start, under 0.01\%. With debug feedback, the ensemble
is the only one of the eight runs that improves the kernel at all; Llama~3.3~70B and Qwen~3.5~122B-A10B both end at
the start. With bare feedback, Qwen~3.5~122B-A10B improves it more (30.5\%), and Llama~3.3~70B again does not improve
it, although every one of its programs builds a correct kernel.

\section{Related work}
\label{sec:related}

\paragraph{Self-improvement.} The idea goes back to \citet{good1966speculations} and the G\"odel
machine~\citep{schmidhuber2003godel}; we study only its weak, inference-time sense, in which no weights change and the
loop cannot rewrite its evaluator, search rule or templates. Recent language-model systems improve scaffolds, prompts,
agent designs or whole research
projects~\citep{zelikman2023stop,fernando2023promptbreeder,hu2024adas,zhang2025dgm,agrawal2025gepa,lu2024aiscientist}.
Most report their strongest results with frontier models. They change what the model is asked to do; we change
\emph{who} proposes, and ask whether several small models can do the work of one large one. Our three components act
on the loop around the models and leave the improved object unchanged, so they could be combined with any of these
systems.

\paragraph{Evolutionary code search with several models.} Language models paired with an evaluator and a program
archive now drive code
search~\citep{romeraparedes2024funsearch,novikov2025alphaevolve,sharma2025openevolve,lange2025shinkaevolve,codeevolve2025};
those that mix models choose one per proposal before it writes. The choice follows a fixed weight or a bandit that
learns from past rewards~\citep{lange2025shinkaevolve}, and later parents are drawn from a shared archive. A weak model
thus still uses calls, and one model's idea reaches the others only when its program is later drawn as a parent.
Closest to our motivation, \citet{donaghy2026dei} find that a mix of frontier models beats a single model at equal
budget. In \method{}, the models meet every round: all write from the same parent, the score chooses among their
programs, and they share one memory and failure log, so one model's idea becomes the next parent of all. We also use
only small open-weight models and compare the ensemble with each member alone and with larger models of the same
families. Simple baselines can rival code evolution~\citep{simplebaselines2026}, so every comparison here fixes the
start program and budget. Published results also differ in model, budget and evaluator, so we run every comparison
inside one loop rather than against reported numbers.

\paragraph{Exploration in autoresearch.} To escape stalls, one line of work samples more varied proposals from one
model~\citep{zhang2025verbalized}; another resets the conversation and keeps only the workspace or the best
candidate~\citep{cai2026forkflush,tan2026forgetting}. Stalls come partly from context inertia: an
agent keeps building on its own earlier attempts~\citep{breunig2026inertia}. Both remedies keep a single model, so the
new proposals still come from one model's training and habits. They diversify one model's proposals; we diversify the
models, in every round rather than after a stall. Our results bear on the claim that language models are strikingly
alike~\citep{jiang2025hivemind}: in ML engineering, the best of five small models changes from task to task. If the
small models were interchangeable, an ensemble would only add samples; on the kernel task, the members alone write more
programs than the ensemble and still gain nothing (\cref{sec:kernel}).

\paragraph{ML-engineering agents.} On MLE-bench~\citep{chan2024mlebench}, AIDE, AIRA-Dojo, ML-Master, MLE-STAR and
MLEvolve~\citep{jiang2025aide,toledo2025airadojo,liu2025mlmaster,nam2025mlestar,mlevolve2026} report results with
frontier models and up to a day per competition. They differ in how they search, from tree search over draft
solutions~\citep{jiang2025aide} to targeted refinement of single code blocks~\citep{nam2025mlestar}. We study the
opposite end and report cost with every result. Our runs use one CPU laptop and \$1.42 per competition for the ensemble
(\cref{sec:cost}); \cref{tab:leaderboard} lists the leaderboard agents beside them.

\paragraph{Ensembles and test-time compute.} Ensembles combine models or samples for one query by fusion, debate or
vote~\citep{wang2024moa,jiang2023llmblender,du2023debate,li2024moreagents,wang2023selfconsistency}. Mixture-of-Agents,
the closest in form, passes every model's answer to the next layer of models~\citep{wang2024moa}. Repeated sampling and
compute-optimal scaling trade inference for quality~\citep{brown2024monkeys,snell2024scaling}: sampling pays off most
when an automatic verifier can pick the right sample~\citep{brown2024monkeys}, and extra test-time compute can let a
smaller model beat a larger one~\citep{snell2024scaling}. Our evaluator plays the verifier, and we see the same effect
across models rather than within one. Routing picks one model per query to save
cost~\citep{chen2023frugalgpt,ong2024routellm}; we call every model in every round and keep the cost low by using small
ones. We ensemble inside a long improvement loop with a shared memory, where one member's program becomes every
member's next parent, and choose by a measured score rather than a vote, a fusion or a judge.




\bibliography{references}
\bibliographystyle{icml2026}

\clearpage
\appendix
\onecolumn

\section{Supplementary results}

\begin{table}[h]
\centering\small
\caption{\textbf{MLE-bench Lite, for context.} Top: any-medal rates on the Lite (low-complexity) split from the
official MLE-bench leaderboard~\citep{chan2024mlebench}, with each agent's model and running time per competition;
frontier models and 12--24 hours, so not comparable with our runs. Bottom: our runs over the \nLite{} Lite
competitions, CPU only.}

\vspace{6pt}
\begin{tabular}{lrrrrr}
\toprule
Run (ours, CPU only) & Any medal (\%) & Valid submission (\%) & $\geq$Median (\%) & Gold (\%) & API cost / comp. \\
\midrule
\textbf{Ensemble (5 small)} & \textbf{18.2} & \textbf{95.5} & \textbf{40.9} & \textbf{4.5} & \textbf{\$1.42} \\
Qwen 3.5 9B & 13.6 & 90.9 & 22.7 & 4.5 & \$0.41 \\
Ministral 8B & 13.6 & 95.5 & 18.2 & 4.5 & \$0.39 \\
Granite 4.2 8B & 13.6 & 100.0 & 13.6 & 4.5 & \$0.25 \\
Gemma 3 12B & 9.1 & 100.0 & 9.1 & 4.5 & \$0.24 \\
Llama 3.1 8B & 9.1 & 95.5 & 9.1 & 4.5 & \$0.39 \\
Llama 3.3 70B & 9.1 & 100.0 & 9.1 & 4.5 & \$0.58 \\
Qwen 3.5 122B-A10B & 27.3 & 95.5 & 40.9 & 18.2 & \$1.90 \\
\bottomrule
\end{tabular}

\end{table}

\begin{table}[h]
\centering\small
\caption{\textbf{Cost per improvement.} RSI cost: the run's own calls (shared zero-shot rounds excluded). Jumps: how
often the returned program changed. Cost per test point: RSI cost per competition over the mean improvement in
percentage points. Stalled: share of rounds inside a stretch of at least 20 without a jump.}
\label{tab:cost-per-gain}
\resizebox{\textwidth}{!}{\begin{tabular}{lrrrrrr}
\toprule
Run & RSI cost / comp. & Jumps / comp. & Jumps / 100 programs & Cost / jump & Cost / test point & Stalled \\
\midrule
\textbf{Ensemble (5 small)} & \textbf{\$1.34} & \textbf{14.1} & \textbf{1.4} & \textbf{\$0.095} & \textbf{\$0.035} & \textbf{66} \\
\midrule
Qwen 3.5 9B & \$0.33 & 5.4 & 2.7 & \$0.062 & \$0.012 & 86 \\
Ministral 8B & \$0.31 & 0.8 & 0.4 & \$0.404 & \$0.040 & 99 \\
Granite 4.2 8B & \$0.18 & 4.6 & 2.3 & \$0.039 & \$0.013 & 88 \\
Gemma 3 12B & \$0.16 & 1.5 & 0.8 & \$0.105 & \$0.046 & 97 \\
Llama 3.1 8B & \$0.31 & 2.4 & 1.2 & \$0.129 & \$0.038 & 95 \\
\midrule
Llama 3.3 70B & \$0.50 & 2.8 & 1.4 & \$0.181 & \$0.040 & 95 \\
Qwen 3.5 122B-A10B & \$1.83 & 9.0 & 4.5 & \$0.204 & \$0.044 & 80 \\
\bottomrule
\end{tabular}
}
\end{table}

\begin{table}[h]
\centering\small
\caption{\textbf{Members inside the ensemble} over \nLite{} competitions. Programs and new bests are totals; tokens
and cost are per competition. New bests: programs that became the ensemble's new best. Final program: competitions
where the member wrote the program the ensemble returned. Best alone: competitions where the member's own run ended best
among the five members alone.}
\label{tab:ideas}
\begin{tabular}{lrrrrrrr}
\toprule
 & \multicolumn{5}{c}{Inside the ensemble} & \multicolumn{2}{c}{Per competition} \\
\cmidrule(lr){2-6}\cmidrule(lr){7-8}
Member & Programs & Valid & New bests & Final program & Best alone & Tokens (M) & Cost \\
\midrule
Qwen 3.5 9B & 4,400 & 19 & 43 (14) & 3 of 22 & 12 of 22 & 2.54 & \$0.32 \\
Ministral 8B & 4,400 & 9 & 26 (8) & 1 of 22 & 3 of 22 & 2.50 & \$0.27 \\
Granite 4.2 8B & 4,400 & 49 & 80 (26) & 4 of 22 & 3 of 22 & 2.53 & \$0.19 \\
Gemma 3 12B & 4,400 & 51 & 82 (26) & 5 of 22 & 1 of 22 & 2.65 & \$0.19 \\
Llama 3.1 8B & 4,400 & 38 & 79 (25) & 9 of 22 & 3 of 22 & 2.21 & \$0.37 \\
\bottomrule
\end{tabular}

\end{table}

\section{Every competition}
\label{app:competitions}

\begin{table}[h]
\centering\small
\caption{\textbf{The competitions:} metric and direction; Kaggle teams on MLE-bench's leaderboard copy; the shared
start program's test score and whose zero-shot program it is ($^{*}$: no member wrote a runnable one in 30 calls).}
\label{tab:competitions}
\begin{tabular}{llcrrl}
\toprule
Competition & Metric & & Teams & Start (test) & Start from \\
\midrule
Aerial Cactus & auc-roc & $\uparrow$ & 1221 & 0.9739 & Qwen 9B \\
Denoising Documents & root\_mean\_squared\_error & $\downarrow$ & 162 & 0.2029 & Granite 8B \\
Spooky Author & multi-class-log-loss & $\downarrow$ & 1242 & 0.5833 & Gemma 12B \\
Random Acts of Pizza & auc-roc & $\uparrow$ & 462 & 0.5694 & Granite 8B \\
Nomad2018 & mean-column-wise-rmsle & $\downarrow$ & 879 & 0.0640 & Qwen 9B \\
Leaf & multi-class-log-loss & $\downarrow$ & 1596 & 0.9310 & Gemma 12B \\
TPS May 2022 & auc-roc & $\uparrow$ & 1152 & 0.8826 & Qwen 122B$^{*}$ \\
Jigsaw Toxic Comments & column-wise ROC AUC & $\uparrow$ & 4539 & 0.9592 & Granite 8B \\
MLSP 2013 Birds & auc-roc & $\uparrow$ & 81 & 0.6764 & Qwen 122B$^{*}$ \\
Right Whale Redux & auc-roc & $\uparrow$ & 129 & 0.5476 & Granite 8B \\
Detecting Insults & auc-roc & $\uparrow$ & 50 & 0.8806 & Gemma 12B \\
Text Normalization (en) & accuracy & $\uparrow$ & 261 & 0.9900 & Granite 8B \\
Plant Pathology 2020 & mean-column-wise-roc-auc & $\uparrow$ & 1318 & 0.5472 & Granite 8B \\
Text Normalization (ru) & accuracy & $\uparrow$ & 163 & 0.9735 & Qwen 9B \\
TPS Dec 2021 & multi-class-classification-accuracy & $\uparrow$ & 1189 & 0.9094 & Gemma 12B \\
Dog Breed & multi-class-log-loss & $\downarrow$ & 1281 & 4.7828 & Gemma 12B \\
Dogs vs. Cats & log-loss & $\downarrow$ & 1315 & 0.5664 & Qwen 9B \\
NYC Taxi Fare & rmse & $\downarrow$ & 1485 & 10.0293 & sample file$^{*}$ \\
Histopathologic Cancer & auc-roc & $\uparrow$ & 1149 & 0.8650 & Qwen 9B \\
RANZCR CLiP & auc-roc & $\uparrow$ & 1547 & 0.5171 & Gemma 12B \\
APTOS 2019 & quadratic-weighted-kappa & $\uparrow$ & 2929 & 0.7395 & Qwen 9B \\
SIIM-ISIC Melanoma & auc-roc & $\uparrow$ & 3308 & 0.6482 & Gemma 12B \\
Spaceship Titanic & accuracy & $\uparrow$ & 2288 & 0.7954 & Granite 8B \\
\bottomrule
\end{tabular}

\end{table}

\begin{table}[h]
\centering\small
\caption{\textbf{Improvement over the start on the test set, per competition and run} (\%). Superscripts:
MLE-bench's grade ($G$ gold, $S$ silver, $B$ bronze, $M$ above the median team). ``--'': no test score. On
\StartMedals{} competitions the shared start alone already medals.}
\label{tab:per-competition}
\begin{tabular}{lrrrrrrrr}
\toprule
Competition & \rotatebox{70}{Ensemble} & \rotatebox{70}{Qwen 9B} & \rotatebox{70}{Ministral 8B} & \rotatebox{70}{Granite 8B} & \rotatebox{70}{Gemma 12B} & \rotatebox{70}{Llama 8B} & \rotatebox{70}{Llama 70B} & \rotatebox{70}{Qwen 122B} \\
\midrule
Aerial Cactus & 97.2$^{M}$ & 98.3$^{M}$ & 60.5 & 70.2 & 3.4 & 49.0 & 66.0 & 98.5$^{M}$ \\
Denoising Documents & 70.5$^{M}$ & 8.1 & 64.1$^{M}$ & 24.9 & 0.0 & 24.9 & 0.0 & 93.0$^{G}$ \\
Spooky Author & 39.3$^{M}$ & 35.8$^{M}$ & 0.0 & 16.8 & 12.9 & 19.1 & 0.0 & 33.6$^{M}$ \\
Random Acts of Pizza & 36.8$^{B}$ & $-$3.6 & 32.7$^{B}$ & $-$1.5 & $-$9.7 & 0.0 & 1.8 & 0.0 \\
Nomad2018 & 4.8$^{S}$ & 5.1$^{S}$ & 0.0$^{B}$ & 3.5$^{S}$ & 2.6$^{B}$ & 7.1$^{S}$ & 4.9$^{S}$ & 4.9$^{S}$ \\
Leaf & 94.5$^{M}$ & 78.7 & $-$79.7 & 58.3 & 0.0 & 15.1 & 30.2 & 21.2 \\
TPS May 2022 & 54.9 & 50.3 & 0.0 & 17.2 & 0.0 & 13.5 & 0.0 & 82.2$^{M}$ \\
Jigsaw Toxic Comments & 51.5 & 45.5 & 27.5 & 7.7 & 27.1 & 12.1 & 8.9 & 52.0 \\
MLSP 2013 Birds & 41.6 & 43.9 & 0.0 & 0.0 & 0.0 & $-$2.4 & 41.9 & 48.7 \\
Right Whale Redux & 5.2 & 2.3 & $-$0.6 & 18.8 & 3.3 & 0.0 & 4.9 & 11.2 \\
Detecting Insults & 23.3$^{G}$ & 7.7$^{G}$ & 13.4$^{G}$ & 7.1$^{G}$ & 2.5$^{G}$ & 3.0$^{G}$ & 10.3$^{G}$ & 27.1$^{G}$ \\
Text Normalization (en) & 11.5$^{B}$ & 16.3$^{S}$ & -- & 1.6 & 0.0 & 1.6 & 0.0 & 10.1$^{B}$ \\
Plant Pathology 2020 & 34.9 & 49.4 & 43.4 & 8.8 & 7.1 & $-$1.0 & 5.3 & 76.9 \\
Text Normalization (ru) & -- & -- & 0.0 & 20.8$^{B}$ & 0.0 & 0.0 & 0.0 & -- \\
TPS Dec 2021 & 40.3 & 0.0 & 0.0 & 0.0 & 0.0 & 10.7 & 23.7 & 57.4$^{G}$ \\
Dog Breed & 9.4 & 3.5 & 2.2 & 2.8 & 1.8 & 0.0 & 0.0 & 11.9 \\
Dogs vs. Cats & 1.8 & 5.0 & 0.0 & $-$4.1 & $-$1.6 & $-$3.2 & $-$5.8 & 11.4 \\
NYC Taxi Fare & 59.6 & 53.4 & 0.0 & 0.0 & 0.0 & 0.0 & 56.1 & 43.9 \\
Histopathologic Cancer & 68.4$^{M}$ & -- & 0.0 & 0.0 & 3.6 & -- & 0.1 & 89.0$^{G}$ \\
RANZCR CLiP & 37.3 & 47.6 & 0.0 & 39.3 & 24.8 & 19.5 & 24.8 & 51.0 \\
APTOS 2019 & 11.0 & 2.6 & 0.0 & 2.7 & 0.0 & 4.6 & 0.0 & 16.9 \\
SIIM-ISIC Melanoma & 6.0 & 2.4 & 0.0 & 7.3 & 0.0 & 0.0 & 0.0 & 27.8 \\
\midrule
Spaceship Titanic$^{\dagger}$ & 5.1$^{M}$ & 6.2$^{M}$ & 9.0$^{B}$ & 0.6$^{M}$ & $-$2.2 & 3.9$^{M}$ & 0.6$^{M}$ & 10.1$^{S}$ \\
\bottomrule
\end{tabular}

\end{table}

\clearpage
\section{What the programs contain}
\label{app:qualitative}

\begin{figure}[h]
  \centering
  \includegraphics[width=0.8\textwidth]{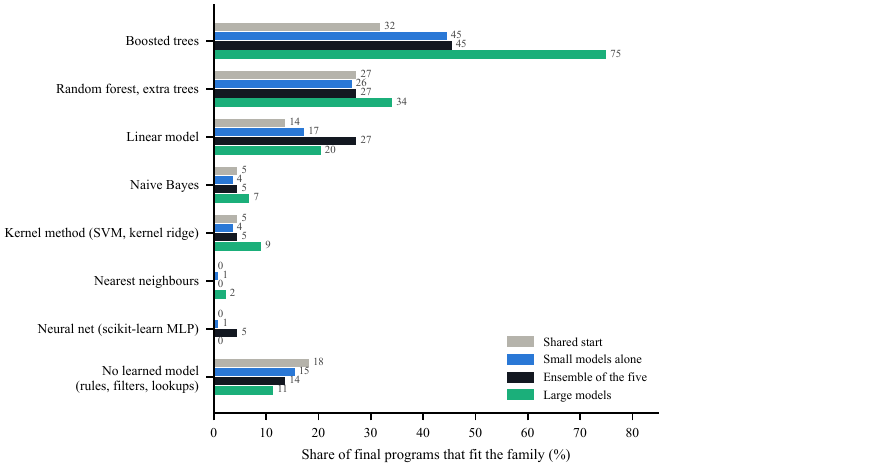}
  \caption{\textbf{Model families in the final programs:} share of final programs whose code fits each family (a
  program can fit several). Small models alone: \QualSmallN{} programs (five models, \nLite{} competitions); ensemble:
  \nLite; large models: \QualLargeN. Grey: the shared start.}
  \label{fig:families}
\end{figure}

This appendix looks inside the programs behind the MLE-bench results. Model families are read from the code of each
final program, with comments, docstrings and imports removed, so a model that is only named does not count
(\cref{fig:families,tab:final-programs}). The ensemble improved on the shared start in all \nLite{} competitions,
against \QualSmallImproved\% of the small models' runs alone and \QualLargeImproved\% of the large models' runs. Its
final programs combine more model families than the small models' alone (\QualEnsFamilies{} per program against
\QualSmallFamilies). Like the large models, it turned many starts into boosted trees on hand-made features:
\QualEnsTrees\% of its final programs fit boosted trees, against \QualStartTrees\% of the starts; the large models did
so most often (\QualLargeTrees\%).

\begin{table}[h]
\centering\small
\caption{\textbf{The final programs by kind of model,} over \nLite{} competitions.}
\label{tab:final-programs}
\begin{tabular}{lrrr}
\toprule
 & Small alone & Ensemble & Large \\
\midrule
Final program is still the start (\%) & 30 & 0 & 18 \\
Model families per final program (mean; start 0.82) & 0.97 & 1.14 & 1.48 \\
Two or more model families (\%) & 9 & 18 & 36 \\
Length of the final program (median characters) & 5,846 & 6,822 & 7,522 \\
\bottomrule
\end{tabular}

\end{table}

\begin{figure}[h]
  \centering
  \includegraphics[width=\textwidth]{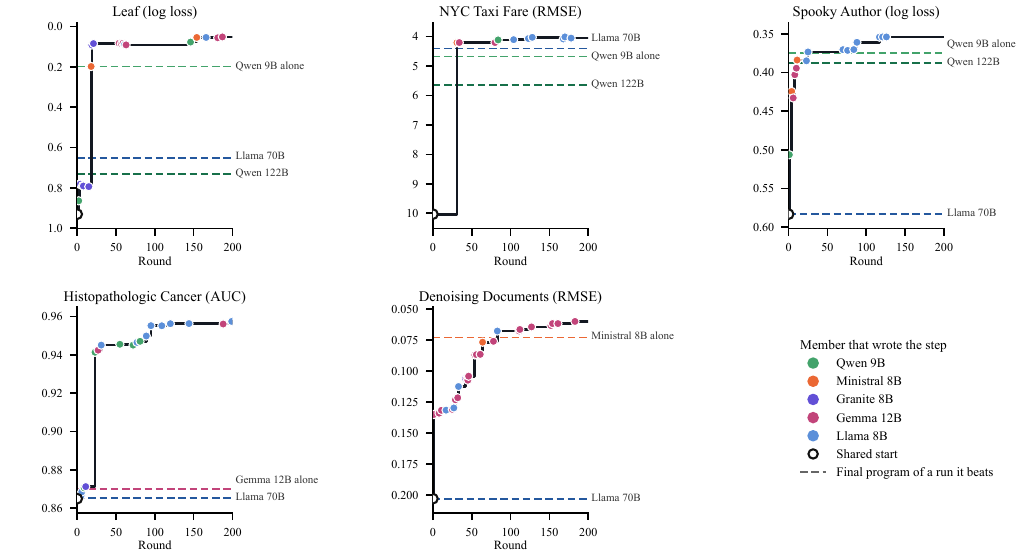}
  \caption{\textbf{How the ensemble's standout programs were built:} test score of each new best over the 200 rounds,
  with each dot coloured by the member that wrote it. Axes are flipped where lower is better, so up is always better.
  Dashed lines: the final programs of the runs it beats.}
  \label{fig:standouts}
\end{figure}

\paragraph{Standout programs.} Below are the final programs that stand out on the test set, each described from its
code. Five are the ensemble's: three that are the best of all eight runs (Leaf, NYC Taxi Fare, Spooky Author) and two
far ahead of every small model alone (Histopathologic Cancer, Denoising Documents). Each was built in steps by several
members (\cref{fig:standouts}): one member's idea makes the large step, and others extend and tune it. No member
reached these programs alone, which illustrates how the ensemble gains over its members (\cref{sec:members}). Three are
Qwen~3.5~122B's gold medals, and one is a small model's best-of-eight silver. Relative error is as in the main text,
and each program is compared with the runs it beats.

\subsubsection*{The ensemble}
\noindent\textbf{Leaf} (above median). Test log loss 0.931 $\to$ 0.05107, the best of the eight runs; relative error 74\% lower than the best small model alone (Qwen 9B), 92\% lower than Llama 70B and 93\% lower than Qwen 122B. \emph{Program:} Uses the 192 shape, margin and texture descriptors the competition provides and adds three summaries of them (mean margin, spread of the shape values, entropy of the texture histogram). A 900-tree balanced random forest and a logistic regression vote on soft probabilities inside isotonic calibration (three folds), and the probabilities are clipped and renormalised. \emph{Path:} Granite made the start's random forest class-balanced and tuned it (iterations 3 to 15). Ministral replaced it with a vote of isotonically calibrated forests at iteration 18, the run's biggest step, and Granite and Gemma tuned the calibrated forest. Ministral added the logistic regression and the summary features at iteration 154, the second big step; Llama 8B and Gemma tuned the result.\par\smallskip
\noindent\textbf{NYC Taxi Fare}. Test RMSE 10.03 $\to$ 4.049, the best of the eight runs; relative error 13\% lower than the best small model alone (Qwen 9B), 8\% lower than Llama 70B and 28\% lower than Qwen 122B. \emph{Program:} Reads the first three million rides and trains LightGBM (1,200 rounds, 120 leaves) on the coordinates, the time (hour, weekday, seconds since 1970), Euclidean and Manhattan distances, the coordinate differences and passenger interactions; no fare is predicted below \$2.50. \emph{Path:} The start only copies the sample submission. Ministral wrote the first real model at iteration 31, LightGBM on two million rides with distance and time features; Gemma raised the training data to five million rides at iteration 34; Gemma, Qwen 9B and then Llama 8B tuned the features and the model from iteration 80 on.\par\smallskip
\noindent\textbf{Spooky Author} (above median). Test log loss 0.5833 $\to$ 0.354, the best of the eight runs; relative error 5\% lower than the best small model alone (Qwen 9B), 39\% lower than Llama 70B and 9\% lower than Qwen 122B. \emph{Program:} Puts character 1--3-gram and word 1--2-gram TF-IDF side by side and stacks three models on them with five-fold cross-validation: a class-balanced logistic regression, a multinomial naive Bayes and a LightGBM, with a small LightGBM on top. \emph{Path:} Qwen 9B paired naive Bayes with a calibrated logistic regression at iteration 1. Ministral added character n-grams and a soft vote at iteration 4, the biggest step. Gemma tuned the vote (iterations 6 to 10), Ministral turned it into a stack at iteration 11, and Llama 8B tuned the stack over nine more new bests.\par\smallskip
\noindent\textbf{Histopathologic Cancer} (above median). Test AUC 0.865 $\to$ 0.9574, second of the eight runs; relative error 67\% lower than the best small model alone (Gemma 12B) and 68\% lower than Llama 70B. \emph{Program:} Describes each tissue patch by sixteen colour and texture statistics at full resolution (per-channel means and spreads, correlations between the channels, brightness, Sobel edge density, contrast and a Laplacian response) and averages a five-fold LightGBM. \emph{Path:} For twenty iterations the members tuned the start's LightGBM on grey pixels. At iteration 23 Qwen 9B replaced the pixels with colour statistics, the big step; Llama 8B, Qwen 9B and Gemma then refined the statistics and the cross-validated model to the end.\par\smallskip
\noindent\textbf{Denoising Documents} (above median). Test RMSE 0.2029 $\to$ 0.05986, second of the eight runs; relative error 18\% lower than the best small model alone (Ministral 8B) and 71\% lower than Llama 70B. \emph{Program:} Learns a lookup table from the clean training pages: each 3$\times$3 window is reduced to a pattern of three levels relative to its local mean and spread, and replaced by the average clean value seen for that pattern. A smoothing filter covers unseen patterns, and edges are blended with the input. \emph{Path:} Gemma and Llama 8B first tuned the start's smoothing filter. Granite's filter learned from the clean pages (iteration 42) and Ministral's multi-scale pattern matching (iteration 64) made the big steps; Gemma refined the result to the end.\par\smallskip
\subsubsection*{Qwen~3.5~122B-A10B}
\noindent\textbf{Histopathologic Cancer} (gold). Test AUC 0.865 $\to$ 0.9851, the best of the eight runs; relative error 89\% lower than the best small model alone (Gemma 12B), 65\% lower than Ensemble and 89\% lower than Llama 70B. \emph{Program:} Shrinks each 96$\times$96 patch to 32$\times$32 pixels and converts it to HSV. It computes histograms of hue, saturation, brightness and gradient magnitude, channel statistics and correlations, local variance at two scales and statistics over a 4$\times$4 grid of cells, and trains one LightGBM (180 leaves, early stopping). There is no convolutional network: every piece is classical image analysis. \emph{Path:} At iteration 2 it replaced the start's grey pixels with colour histograms, its biggest step. It then added HSV colour (iteration 4), gradient features (iteration 7) and statistics over a grid of cells (iteration 74), and kept refining them to iteration 134.\par\smallskip
\noindent\textbf{Denoising Documents} (gold). Test RMSE 0.2029 $\to$ 0.0143, the best of the eight runs; relative error 80\% lower than the best small model alone (Ministral 8B), 76\% lower than Ensemble and 93\% lower than Llama 70B. \emph{Program:} Learns from the noisy and clean training pairs pixel by pixel. Each pixel gets 23 features: its 3$\times$3 neighbourhood, its differences from box, Gaussian and median filters, a difference of Gaussians, how far it sits from the local minimum and maximum, gradients in four directions, the local spread and the page's mean. LightGBM is trained on two million sampled pixels. \emph{Path:} At iteration 5 it replaced the start's smoothing filter with a per-pixel LightGBM, its biggest step. Later new bests added the local minimum and maximum (iteration 29), the Gaussian (iteration 54) and median (iteration 167) filter differences, and finally the gradients and the page's mean (iteration 199).\par\smallskip
\noindent\textbf{TPS Dec 2021} (gold). Test accuracy 0.9094 $\to$ 0.9614, the best of the eight runs; relative error 52\% lower than the best small model alone (Llama 8B), 29\% lower than Ensemble and 44\% lower than Llama 70B. \emph{Program:} Trains LightGBM (600 trees, 40 leaves) on a stratified sample of 550,000 rows with fifteen engineered features: elevation minus the vertical distance to water, counts of active soil and wilderness flags, ratios of the distances to water and roads, total hillshade, and products of elevation, slope and aspect. \emph{Path:} It tuned the start's random forest, switched to LightGBM at iteration 21, sampled 550,000 rows at iteration 49 and added the engineered features at iteration 55; its last new best (iteration 162) tuned them.\par\smallskip
\subsubsection*{Qwen~3.5~9B alone}
\noindent\textbf{Text Normalization (en)} (silver). Test accuracy 0.99 $\to$ 0.9917, the best of the eight runs; relative error 15\% lower than the best small model alone (Granite 8B), 5\% lower than Ensemble, 16\% lower than Llama 70B and 7\% lower than Qwen 122B. \emph{Program:} Memorises spoken forms in context: for each token it takes the most frequent spoken form seen with the same previous and next token, then with the same next token, then for the token alone, and copies tokens it has never seen. \emph{Path:} Starting from a plain lookup of each token, it added the previous token as context at iteration 2, the next token at iteration 70, and the three-level back-off at iterations 170 and 174.\par\smallskip

\section{Kernel study}
\label{app:kernel}

\begin{table}[h]
\centering\small
\caption{\textbf{Kernel optimization.} Final kernel's cycles on the nine-input test (the same as on the two practice
inputs), improvement over the start (\KernStart{} cycles), share of the run's programs that built a correct kernel, and
API cost.}
\label{tab:kernel}
\begin{tabular}{lrrrrrrrr}
\toprule
 & \multicolumn{4}{c}{Bare feedback} & \multicolumn{4}{c}{Debug feedback} \\
\cmidrule(lr){2-5}\cmidrule(lr){6-9}
Run & Cycles & Impr. (\%) & Valid & Cost & Cycles & Impr. (\%) & Valid & Cost \\
\midrule
\textbf{Ensemble (5 small)} & \textbf{131,338} & \textbf{11.1} & \textbf{25} & \textbf{\$1.96} & \textbf{135,445} & \textbf{8.3} & \textbf{25} & \textbf{\$2.28} \\
\midrule
Qwen 3.5 9B & 147,734 & 0.0 & 2 & \$0.61 & 147,734 & 0.0 & 2 & \$0.62 \\
Ministral 8B & 147,734 & 0.0 & 0 & \$0.21 & 147,734 & 0.0 & 0 & \$0.22 \\
Granite 4.2 8B & 147,724 & 0.0 & 2 & \$0.26 & 147,734 & 0.0 & 2 & \$0.27 \\
Gemma 3 12B & 147,732 & 0.0 & 19 & \$0.20 & 147,732 & 0.0 & 16 & \$0.20 \\
Llama 3.1 8B & 147,734 & 0.0 & 10 & \$0.70 & 147,734 & 0.0 & 34 & \$0.72 \\
\midrule
Llama 3.3 70B & 147,734 & 0.0 & 100 & \$0.55 & 147,734 & 0.0 & 85 & \$0.55 \\
Qwen 3.5 122B-A10B & 102,671 & 30.5 & 28 & \$2.16 & 147,734 & 0.0 & 0 & \$2.30 \\
\bottomrule
\end{tabular}

\end{table}

Each program defines \texttt{KernelBuilder.build\_kernel}; the kernel is built in a sandboxed process and simulated in
a separate process that never runs the program's code, on the take-home's frozen problem, with 120\,s of CPU and 2\,GB
per practice evaluation and up to 16,384 output tokens. The prompt contains the take-home's text and problem file, the
interface, the evaluation and the rules, and no published cycle counts. Llama~3.3~70B's two runs were re-run from the
same start after a repair to the notes parser; the first runs ended at the start. Following the take-home's license, we
publish no kernel code.

\section{Implementation, prompt and evaluator}
\label{app:prompt}

We implemented \method{} on the open-source OpenEvolve framework~\citep{sharma2025openevolve}, reusing its prompt
builder, evaluator interface and run bookkeeping: \PatchLines{} lines of code, \ConfigValues{} configuration values and
\TemplateMarkers{} template placeholders, with the framework's own \OELines{} lines unchanged. All models were called
through OpenRouter, each pinned to one provider (precision where the provider states it): Qwen~3.5~9B on Parasail
(bf16), Ministral~8B (2512) on Mistral, Granite~4.2~8B on DeepInfra (bf16), Gemma~3~12B on DeepInfra, Llama~3.1~8B on
Novita (fp8), Llama~3.3~70B on Parasail (fp8) and Qwen~3.5~122B-A10B on Alibaba.

The system message has four versioned, hashed sections (prompt mle-v1): the role (``You are an expert in machine
learning and data science. Design an algorithm for this machine-learning competition.''); the task (MLE-bench's
description, the file listing with sample rows, the \texttt{solve} interface, the evaluation, and the rules: libraries,
no network or subprocesses, deterministic, one core, 300\,s CPU, 4\,GB, at most 4,000 characters of literal data); the
output format and idea card (name, idea, numbered steps); and what the feedback numbers mean. The user message is a
full-rewrite template with the parent and its practice results, the run's recent failed programs with their errors,
the strategy notes, the request to rewrite the program, and last, the request to rewrite the notes in a
\texttt{notes} block. A program without a complete idea card is rejected before it runs.

\end{document}